\documentclass[10pt]{article}
\usepackage[a4paper, total={5in, 8in}]{geometry}

\usepackage{times}
\usepackage{epsfig}

\usepackage[utf8]{inputenc}
\usepackage{graphicx,verbatimbox}
\usepackage{tcolorbox}

\usepackage{rotating}
\usepackage{palatino}

\usepackage{tikz}
\usepackage{comment}
\usepackage{amsmath,amssymb} %

\usepackage{booktabs} %
\usepackage{xcolor}
\usepackage{color,colortbl}
\usepackage{pifont}

\usepackage{multirow}
\usepackage[font=footnotesize]{caption}
\usepackage{placeins}
   
\usepackage{xspace}

\usepackage{booktabs}
\usepackage{multirow}

\usepackage{booktabs}
\usepackage{xspace}
\usepackage{comment}
\usepackage{gensymb}
\usepackage{tabularx}

\makeatletter
\DeclareRobustCommand\onedot{\futurelet\@let@token\@onedot}
\def\@onedot{\ifx\@let@token.\else.\null\fi\xspace}

\usepackage[colorlinks=true,linkcolor=blue, citecolor=cyan]{hyperref}

\def\eg{\emph{e.g}\onedot} 
\def\ie{\emph{i.e}\onedot}

\def\wrt{w.r.t\onedot} 
\def\etal{\emph{et al}\onedot}

\def\ourdataset{WOD: PVPS\xspace}

\newcommand{\figref}[1]{Fig\onedot~\ref{#1}}
\newcommand{\equref}[1]{Eq\onedot~\eqref{#1}}
\newcommand{\secref}[1]{Sec\onedot~\ref{#1}}
\newcommand{\tabref}[1]{Tab\onedot~\ref{#1}}

\usepackage{pifont}
\newcommand{\cmark}{\ding{51}}
\newcommand{\xmark}{\ding{55}}

\usepackage{caption}
\usepackage{subcaption}

\usepackage{microtype}

\newlength\savewidth

\iftrue 
\newcommand{\cutsectionup}{\vspace*{-10pt}}
\newcommand{\cutsectiondown}{\vspace*{-6pt}}
\newcommand{\cutsubsectionup}{\vspace*{-8pt}}
\newcommand{\cutsubsectiondown}{\vspace*{-4pt}}

\newcommand{\cutcaptionup}{\vspace*{-10pt}}
\newcommand{\cutcaptiondown}{\vspace*{-10pt}}
\newcommand{\cuthalftablecaptionup}{\vspace*{-5pt}}
\newcommand{\cuthalftablecaptiondown}{\vspace*{-5pt}}

\else 
\newcommand{\cutsectionup}{}
\newcommand{\cutsectiondown}{}
\newcommand{\cutsubsectionup}{}
\newcommand{\cutsubsectiondown}{}

\newcommand{\cutcaptionup}{}
\newcommand{\cutcaptiondown}{}
\newcommand{\cuthalftablecaptionup}{}
\newcommand{\cuthalftablecaptiondown}{}

\fi

\title{Waymo Open Dataset:\\Panoramic Video Panoptic Segmentation}
\author{
\begin{minipage}{\linewidth}
\begin{center}
\normalsize Jieru Mei$^{1}$\footnote{Work done as an intern at Waymo.} \quad Alex Zihao Zhu$^{2}$ \quad Xinchen Yan$^{2}$ \quad Hang Yan$^{2}$ \\[1ex] Siyuan Qiao$^{3}$ \quad Yukun Zhu$^{3}$ \quad Liang-Chieh Chen$^{3}$ \\[1ex] \quad Henrik Kretzschmar$^{2}$ \quad Dragomir Anguelov$^{2}$ \\[1em]
\scalebox{1.}{$^1$Johns Hopkins University \quad $^2$Waymo LLC \quad $^3$Google Research}
\end{center}
\end{minipage}
}

\date{~}

\begin{document}
\maketitle
\begin{abstract}
Panoptic image segmentation is the computer vision task of finding groups of pixels in an image and assigning semantic classes and object instance identifiers to them.
Research in image segmentation has become increasingly popular due to its critical applications in robotics and autonomous driving.
The research community thereby relies on publicly available benchmark dataset to advance the state-of-the-art in computer vision.
Due to the high costs of densely labeling the images, however, there is a shortage of publicly available ground truth labels that are suitable for panoptic segmentation.
The high labeling costs also make it challenging to extend existing datasets to the video domain and to multi-camera setups.
We therefore present the Waymo Open Dataset: Panoramic Video Panoptic Segmentation Dataset, a large-scale dataset that offers high-quality panoptic segmentation labels for autonomous driving. We generate our dataset using the publicly available Waymo Open Dataset, leveraging the diverse set of camera images.
Our labels are consistent over time for video processing and consistent across multiple cameras mounted on the vehicles for full panoramic scene understanding.
Specifically, we offer labels for 28 semantic categories and 2,860 temporal sequences that were captured by five cameras mounted on autonomous vehicles driving in three different geographical locations, leading to a total of 100k labeled camera images.
To the best of our knowledge, this makes our dataset an order of magnitude larger than existing datasets that offer video panoptic segmentation labels.
We further propose a new benchmark for Panoramic Video Panoptic Segmentation
and establish a number of strong baselines based on the DeepLab family of models.
We will make the benchmark and the code publicly available, which we hope will facilitate future research on holistic scene understanding. Find the dataset at \url{https://waymo.com/open}.
\end{abstract}

\cutsectionup
\section{Introduction}
\cutsectiondown

\begin{figure}[t]
    \centering
    \scalebox{0.95}{
    \includegraphics[width=\textwidth]{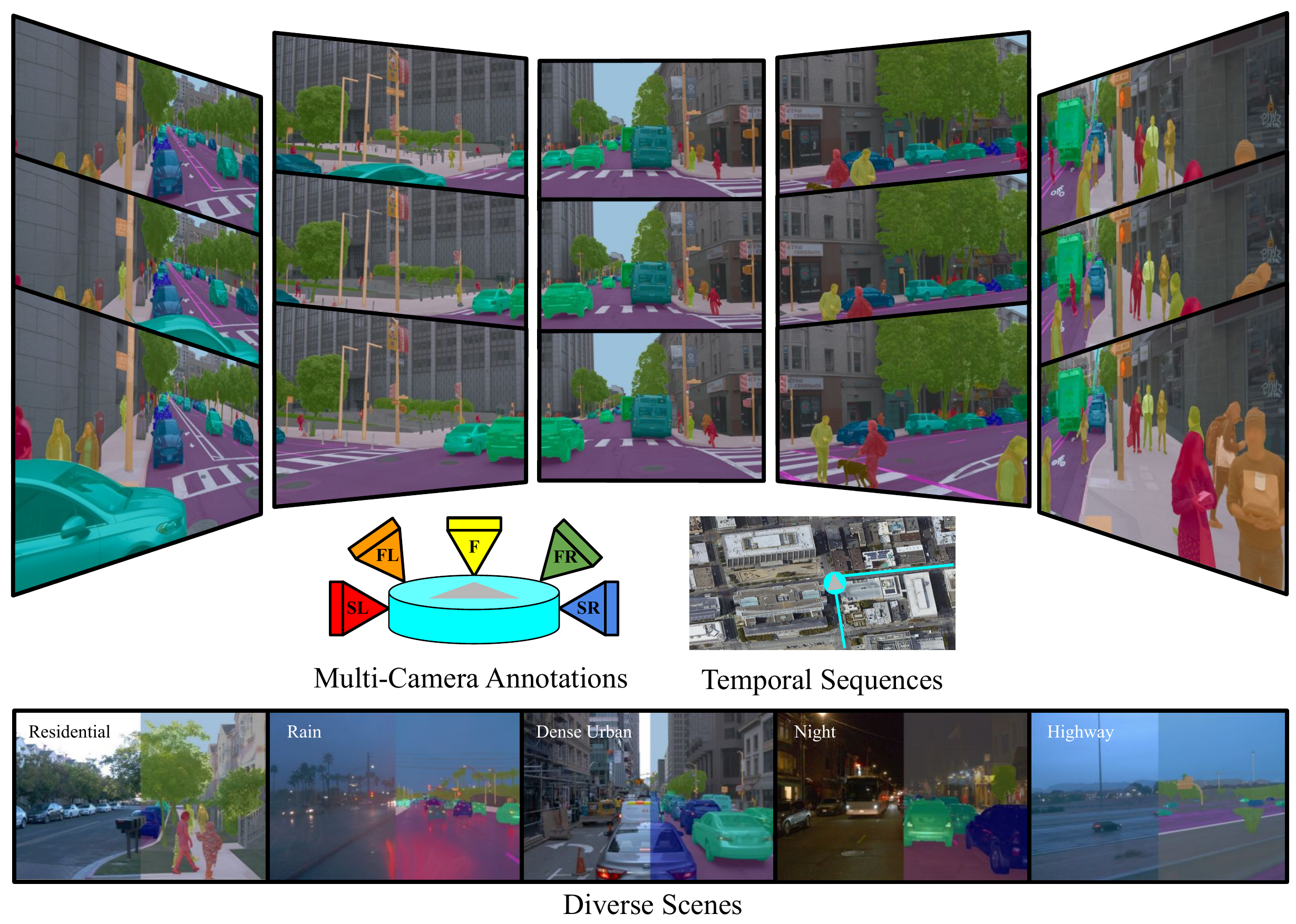}
    }
    \cutcaptionup
    \caption{We provide panoptic segmentation labels for 100k~camera images of the Waymo Open Dataset. Our dataset is grouped into 2,860 temporal sequences captured by five cameras, mounted on autonomous vehicles driving in three geographical locations. Instance segmentation labels are consistent both across cameras and over time. Our dataset offers diversity in terms of object classes, locations, weather, and time of day.
    }
    \label{fig:sample_sequence}
    \cutcaptiondown
\end{figure}

Semantic visual scene understanding has been studied extensively for decades in the field of computer vision~\cite{shi2000normalized,tu2005image,felzenszwalb2004efficient,ladicky2010and,yao2012describing,xu2012streaming}. Researchers have tackled tasks of varying difficulty, ranging from segmenting distinct objects in individual camera images~\cite{He2004CVPR,Hariharan2014ECCV,Long2015CVPR,Chen2018TPAMI} to tracking and segmenting multiple objects in videos~\cite{wu2013online,Voigtlaender19CVPR_mots,dendorfer2020ijcv}.
Robotic applications, such as autonomous driving, have led to new challenges and opportunities for semantic visual scene understanding~\cite{Geiger12CVPR,Cordts16CVPR}.

Modern autonomous vehicles tend to be equipped with multiple cameras and LiDAR scanners.
The cameras provide rich semantic information about the scene, whereas the LiDAR scanners capture sparse, but geometrically highly accurate information.
Autonomous vehicles need to be able to fuse and interpret the data stream from multiple sensors to build and maintain over time an accurate and consistent estimate of the world.
One challenge when tracking and segmenting multiple objects is that 
objects of interest may leave the field of view of a camera to enter the field of view of another camera across consecutive video frames.

In this paper, we study the new task of video panoptic segmentation~\cite{Kirillov19CVPR,Kim20CVPR} for autonomous vehicles equipped with multiple cameras. See \figref{fig:sample_sequence} for an illustration.
Panoptic segmentation enables autonomous vehicles to reason about their surroundings in terms of semantic and geometry properties, such as fine-grained object contours.
There are also important offboard applications, including auto-labeling~\cite{zakharov2020autolabeling,yang2021auto4d,qi2021offboard} and camera sensor simulation~\cite{mallya2020world,ling2020variational,chen2021geosim}.
On the one hand, most existing panoptic segmentation datasets~\cite{Cordts16CVPR,neuhold2017mapillary} provide labels for individual camera images. This makes it difficult to train models that fuse information from multiple camera images, either temporally or by leveraging a multi-camera setup.
On the other hand, datasets that provide panoptic segmentation labels for video data~\cite{Kim20CVPR,Weber2021NEURIPSDATA} tend to be scarce and much smaller than datasets for object detection and tracking for autonomous driving~\cite{Geiger12CVPR,Sun2020CVPR}.
To bridge this gap, we present a new benchmark dataset for panoptic segmentation based on the popular Waymo Open Dataset (WOD).
Specifically, we provide panoptic segmentation labels for video data that are consistent across five cameras mounted on the vehicles.
We further present a benchmark that captures the task of multi-camera panoptic segmentation in video data for autonomous driving.
Overall, we provide panoptic segmentation labels for 100k camera images, which we group into training (70\%), validation (10\%) and test (20\%) sets. The training set consists of 2,800~sequences, each of which comprises labels for five cameras spanning 1.2 seconds and five temporal frames.
In contrast, our validation and test sets consist of 60 longer sequences, in order to facilitate the evaluation of long-term tracking. Each validation and test sequence consists of 100~temporal frames, spanning the full 20s of a scene, while also providing labels across all five cameras.
We extend the Segmentation and Tracking Quality (STQ) metric~\cite{Weber2021NEURIPSDATA} to support our multi-camera setup by computing a weight for pixels depending on the cameras they correspond to.
We also extend a state-of-the-art video panoptic segmentation method, ViP-DeepLab~\cite{Qiao2021CVPR}, to our multi-camera setup by training separate models on each camera view and by training a model on a panorama generated from all views. We present an extensive experimental evaluation on the proposed dataset and metric. 

We published the full dataset to enhance video panoptic segmentation research while also opening up the field of panoramic video panoptic segmentation.

\cutsectionup
\section{Related Work}
\cutsectiondown
{\bf Panoptic Segmentation}\quad
The task of panoptic segmentation~\cite{Kirillov19CVPR} aims to unify semantic segmentation~\cite{He2004CVPR} and instance segmentation~\cite{Hariharan2014ECCV}, requiring assigning a class label and instance ID to all pixels in an image.
Modern panoptic segmentation systems could be roughly categorized into top-down (or proposal-based)~\cite{Kirillov2019CVPR_fpn,Porzi2019CVPR,Li2018CVPR_attention,Liu2019CVPR,Xiong2019CVPR,Weber2020IROS}
and bottom-up (or proposal-free)~\cite{Yang2019Arxiv,Gao2019ICCV,Wang2020CVPR_voting,Cheng20CVPR,Wang2020ECCV_axial} approaches. Our adopted baseline methods belong to the bottom-up category.

{\bf Video Panoptic Segmentation}\quad
Extending panoptic segmentation to the video domain, Video Panoptic Segmentation (VPS)~\cite{Kim20CVPR} requires generating the instance tracking IDs (\ie, temporally consistent instance IDs) along with panoptic segmentation results across video frames. Current VPS datasets are small scale in terms of semantic classes and sizes. Specifically, Cityscapes-VPS~\cite{Kim20CVPR} sparsely annotates (every five frame) Cityscapes~\cite{Cordts16CVPR} video sequences, resulting in only 3,000 frames with 19 semantic classes for training and testing. Recently, STEP~\cite{Weber2021NEURIPSDATA} extends KITTI-MOTS~\cite{Geiger12CVPR,Voigtlaender19CVPR_mots} and MOTS-Challenge~\cite{Voigtlaender19CVPR_mots,dendorfer2020ijcv} for VPS. However, their annotated datasets are still small-scale (18K annotated frames with 19 semantic classes for KITTI-STEP, and 2K frames with 8 classes for MOTChallenge-STEP), and the video sequences are only captured by a single front-view camera. On the other hand, our annotated dataset presents the first large-scale VPS annotations and extends to the multi-camera scenario.

{\bf Segmentation Benchmarks}\quad
There are other popular video segmentation benchmarks existing in the literature, \eg, VSPW~\cite{Miao2021CVPR} for video semantic segmentation, while MOTS~\cite{Voigtlaender19CVPR_mots} and Youtube-VIS~\cite{Yang19ICCV} for video instance segmentation. Our benchmark is also related to urban scene understanding, where typical benchmarks include~\cite{brostow2009semantic,Geiger12CVPR,Lin14ECCV,Cordts16CVPR,neuhold2017mapillary,chang2019argoverse,yogamani2019woodscape,behley2019semantickitti,Sun2020CVPR,caesar2020nuscenes,liang2020polytransform,huang2018apolloscape,yu2020bdd100k,yang2021capturing,liao2021kitti,geyer2020a2d2}.
Our work is most related to WildPASS~\cite{yang2021capturing}, which also aims to endow machines with large field-of-view perception. However, building on top of the large-scale Waymo Open Dataset~\cite{Sun2020CVPR}, our benchmark provides much more high-quality annotated video sequences.

{\bf Multi-Camera Multi-Object Tracking}\quad
Consistently tracking objects across multiple cameras, multi-camera multi-object tracking~\cite{fleuret2007multicamera,eshel2008homography,berclaz2011multiple,roshan2012gmcp,hofmann2013hypergraphs,dehghan2015gmmcp,baque2017deep,ristani2018features} has been a popular research topic in the computer vision community. Typical benchmarks~\cite{ferryman2009pets2009,kuo2010inter,xu2016multi,ristani2016performance,chavdarova2018wildtrack,tang2019cityflow,han2021mmptrack} only track a single class (\eg, people or vehicles) with bounding boxes, while our proposed benchmark demands for pixel-level tracking and segmentation for multiple classes.

{\bf Panoramic Semantic Segmentation}\quad
Panoramic semantic segmentation provides surround-view perception~\cite{song2018im2pano3d,narioka2018understanding,tateno2018distortion,zhang2019orientation,yang2021capturing,yang2019pass}, but limited to semantic segmentation without temporal and instance-level understanding. Our work is similar, but additionally tackles video panoptic segmentation. Recently, ~\cite{roddick2020predicting,philion2020lift} predict bird's-eye view semantic segmentation using multi-camera inputs.

\cutsectionup
\section{\ourdataset Dataset}
\cutsectiondown
In this section, we first recap the existing Waymo Open Dataset (WOD)~\cite{Sun2020CVPR}, one of the largest and most diverse multi-sensor datasets in the autonomous driving domain.
We leverage the existing data that comes with coarse-level annotations (e.g., 2D and 3D bounding boxes) as the foundation, and subsample images for our dataset.
We then provide an overview of our \ourdataset dataset, including panorama generation, statistics of the semantic classes, and temporal frame sampling.
Finally, we explain in details our hybrid scheme to address the challenges in multi-camera and video labeling.
We obtain consistent instance IDs across temporal frames and cameras by associating the panoptic labels from each individual image with the existing box-level annotations.

\begin{table}[!t]
\caption{Dataset comparison. Our \ourdataset is a new large-scale panoramic video panoptic segmentation dataset. $^\dagger$WildPASS contains 500 panoramas.
  }
  \centering
  \scalebox{0.66}{
  \begin{tabular}{l | c | c c c c }
    \toprule
    dataset statistics & \ourdataset (ours) &
    WildPASS~\cite{yang2021capturing} & Cityscapes-VPS~\cite{Kim20CVPR} & KITTI-STEP~\cite{Weber2021NEURIPSDATA} & MOT-STEP~\cite{Weber2021NEURIPSDATA} \\
    \midrule
    \# sequences & 2860 & - & 500   & 50      & 4 \\
    \# images    & 100,000 & 500$^\dagger$ & 3,000 & 19,103 & 2,075 \\
    \# tracking classes          & 8 & -  & 8        & 2          & 1 \\
    \# semantic classes          & 28 & 8 & 19       & 19         & 7 \\
    panoramic & \cmark & \cmark & \xmark & \xmark & \xmark \\
    video panoptic & \cmark & \xmark & \cmark & \cmark & \cmark\\
    \bottomrule
  \end{tabular}
  }
  \label{tab:dataset_stats}
\end{table}

\cutsubsectionup
\subsection{Dataset Overview}
\cutsubsectiondown
The Waymo Open Dataset contains 1,150 scenes, each consisting of 20~seconds of data captured at 10Hz (\ie, 10 frames per second, and thus 200~frames per scene). Each data frame in the dataset includes 3D point clouds from the LiDAR devices, images from five cameras (positioned at Front, Front-Left, Front-Right, Side-Left, and Side-Right), and ground truth 3D and 2D bounding boxes annotated by humans in the LiDAR point clouds and camera images, respectively. Each bounding box contains an ID that is unique to that object across the entirety of each scene. For the LiDAR data, this allows for tracking in the whole scene. For the camera data, these IDs are consistent within each camera's images only.

Built on top of the WOD, Our \ourdataset dataset consists of 100,000 images with panoptic segmentation labels using a prescribed train, validation, and test set split, subsampled from the existing 1.15~million images. 
In \tabref{tab:dataset_stats}, we compare our proposed \ourdataset dataset with the public datasets for video panoptic segmentation. 
Our dataset is the only one that provides panoptic segmentation annotations that are consistent both across multiple cameras and across time. Furthermore, our dataset is much larger both in terms of number of frames and number of semantic classes than existing datasets~\cite{Kim20CVPR,yang2021capturing,Weber2021NEURIPSDATA}.

{\bf Equirectangular Panorama}\quad
\label{sec:panorama_computation}
We reconstruct the equirectangular panorama (220\degree coverage from five cameras) by stitching each individual camera images as an alternative input format to our dataset.
Specifically, we first use the extrinsics and intrinsics from the five cameras provided by WOD to unproject each pixel coordinates to the 3D space.
We then set a virtual camera~\cite{su2017making} located at the geometric mean of all five camera centers and compute the pixel colors by equirectangular projection from the 3D space with bilinear sampling.
For pixels correspond to multiple camera views, we compute the weights based on the distance of each pixel in the panorama to each of the camera views' boundaries.
For panoptic labels, we compute labels in each camera view given the camera parameters of five cameras and the virtual camera using the nearest sampling. Then we use the method in Qiao \etal~\cite{Qiao2021CVPR} to stitch the panorama labels to maintain the view consistency.
Finally, we fused the five panorama labels together based on the correspondences and the distances to the camera view's boundaries.
There are more sophisticated methods~\cite{schonberger2016pixelwise,thrun2006graph} that leverage cross-frame information and the geometry captured from LiDAR sensors to potentially improve panorama generation. We leave this as an open research topic in the future.

\begin{figure}[!t]
    \centering
    \includegraphics[width=0.95\textwidth]{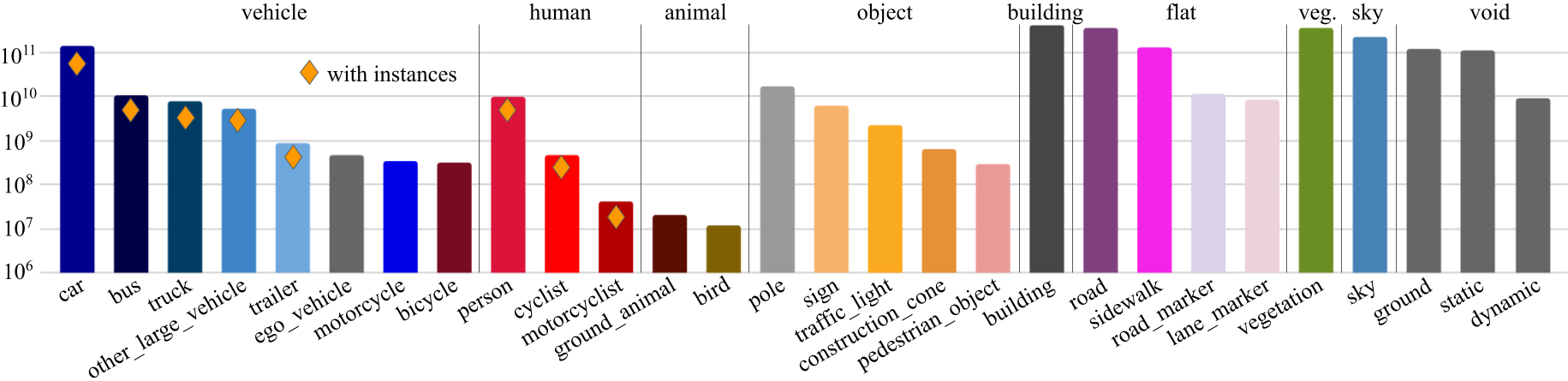}
    \cutcaptionup
    \caption{Histogram of the 28~semantic categories in our dataset in terms of their pixel distributions. The vertical axis denotes the number of pixels for each class in log scale. We provide instance IDs for classes marked with diamonds.}
    \label{fig:class_dist}
    \cutcaptiondown
\end{figure}

\begin{figure}[!t]
    \centering
    \includegraphics[width=0.95\textwidth]{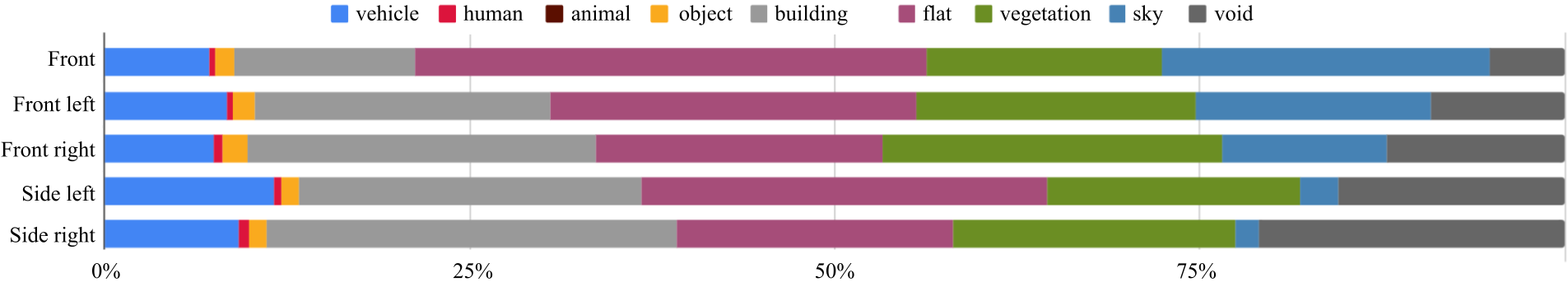}
    \cutcaptionup
    \caption{Super-class distributions for each camera.
    Each camera sees a different distribution of classes, due to their fixed positions and different field-of-views.
    }
    \label{fig:class_dist_per_cam}
    \cutcaptiondown
\end{figure}

{\bf Semantic Class Distribution}\quad
In total, our dataset contains 28 semantic categories, outlined with their frequency in pixels in Fig.~\ref{fig:class_dist}. 
In addition, we provide instance IDs for most of the classes under the \texttt{vehicle} and \texttt{human} super-classes, as they are major dynamic categories in the autonomous driving space.
We also outline the pixel distribution for each camera view in Fig.~\ref{fig:class_dist_per_cam}, where we see notable differences in the distributions in each camera.
For example, the front camera covers more of \texttt{flat} (e.g., road surfaces) and \texttt{sky} pixels  than the rest of the cameras, while the side left camera covers more \texttt{vehicle} pixels due to the ego-vehicle driving on the right hand side of the road.
This analysis is important as machine learning models trained on the images captured by a single camera from the existing datasets may not necessarily generalize to the other cameras due to large domain gaps across different cameras.
In contrast, our proposed task has an emphasis on the holistic scene understanding, which grants our \ourdataset dataset unique value to the research community.

{\bf Temporal Frame Sampling for Human Annotations}\quad
To maximize the diversity of the images on the training set, we subsample \textit{sparsely} from each scene, labeling chunks of five-frame sequences from all the cameras.
We start by randomly selecting 700 out of the 798 scenes. For each scene, which typically has 200 frames, we annotate \textit{four} sets of five-frame sequences, starting at frame indices $\{25, 50, 125, 150\}$ (\ie, we pick 25th, 50th, 125th, and 150th frames as the first frame of each five-frame sequence for annotation). For each set, we further select frames with offsets $\{0, 4, 6, 8, 12\}$ \wrt the first frame for annotations. For example, the first set of five-frame sequences will contain frames with indices $\{25, 29, 31, 33, 37\}$. 
Our \textit{sparse} sampling strategy facilitates a variety of different sequence lengths, allowing users to train on frame pairs with time difference as small as two frames (0.2 seconds) and as large as 12 frames (1.2 seconds). As a result, our training set contains groups of five temporal frames across all five cameras, yielding 2,800 sequences of 25 images (5 temporal frames $\times$ 5 cameras), or 70,000 images in total. Finally, we provide the associations between each instance ID and the corresponding 3D LiDAR bounding box, allowing us to compute very long associations (up to 13.7 seconds between all four sequences), if an object persists across multiple sequences in the same scene.

For the validation and test sets, we aim to enable the testing of long-term consistency across cameras and frames.
We therefore \textit{densely} sample frames at 5Hz from chunks of 100~frames across all cameras (\ie, every other two frames are sampled in the 200~frame sequence).
We select 20 and 40 scenes for validation and test sets by maintaining diversity in the location, density of object, and time of day distributions of WOD. In contrast to the training set, for each scene selected from the validation and test tests, we \textit{densely} subsample the scenes for these splits by labeling every other frame, resulting in sequences with 100 temporal frames across all five cameras. In the end, our validation set contains annotations for 20~sequences of 500~images (100 temporal frames $\times$ 5 cameras), and our test set consists of 40~sequences of 500~images (or totally 10,000 and 20,000 annotations for validation and test sets, respectively). The test set annotations will not be made publicly available, but instead we will prepare a test server to evaluate the held-out test set, once the dataset is released.

\cutsubsectionup
\subsection{Associating Instance IDs Across Cameras and Frames}
\label{sec:temporal_consistency_dataset}
\cutsubsectiondown

\begin{figure}[!t]
    \centering
    \includegraphics[width=0.95\textwidth]{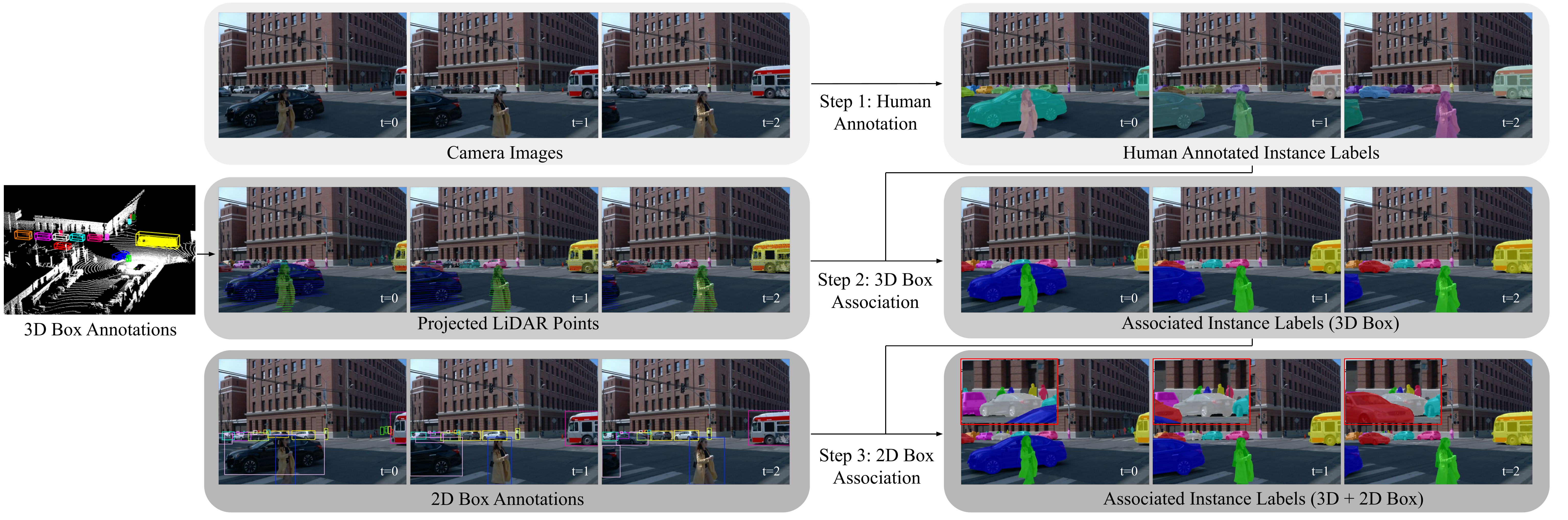}
    \cutcaptionup
    \caption{Labeling and Association Overview. Human annotators first label each camera image for panoptic segmentation separately (step 1).
    LiDAR points within each ground truth 3D bounding box are then projected to each image, and associate with the single frame instance labels (step 2). For far-range instances without corresponding 3D bounding boxes, we associate the single frame instance labels over time using the ground truth 2D bounding boxes within each camera (step 3).
    New associations are highlighted in the zoomed-in views at the bottom.
    }
    \label{fig:lidar_projection}
    \cutcaptiondown
\end{figure}
In constructing the panoramic video panoptic segmentation dataset, ensuring the annotations have consistent instance IDs across cameras and temporal frames is one of the major challenges.
Manual labeling is a straight-forward option, but is time-consuming and expensive at large scales.
In addition, it is difficult to develop an effective labeling interface that allows human annotators to iteratively refine instance labels across cameras and temporary frames.

We instead assigned human annotators to label each camera image for panoptic segmentation separately and employed a hybrid scheme that leverages the existing coarse-level annotations in WOD.
The coarse-level annotations include (i) 3D bounding boxes with corresponding IDs that are consistent across all frames and cameras; and (ii) 2D bounding boxes with IDs that are consistent across temporaral frames, but annotated independently for each camera. 
Associations were then computed between each instance and its corresponding 3D LiDAR boxes
and 2D camera boxes.
Instances determined to correspond to the same object are then mapped to the same ID in all frames across cameras. A sample sequence from this process can be found in Fig.~\ref{fig:lidar_projection}.

For a given frame with instance labels, 3D point clouds, and 3D bounding boxes, we associate instances with boxes by filtering the LiDAR points within each box, and projecting them onto the image. Association scores are then computed using IoU between the convex hull of the projected LiDAR points and each instance label. Bipartite matching is then applied to match each projected box with an instance label.
For 3D driving scenes, points inside the bounding box almost entirely correspond to the instances inside of them, and so these projected LiDAR points have a high overlap with their corresponding instance masks in the image.
Our label association step is related to the prior work~\cite{huang2018apolloscape,liao2021kitti}, but,
our association leverages the ground-truth labeled 3D boxes and only transfers instance IDs rather than fine-grained per-pixel labels.

There are, however, a small number of instances without corresponding LiDAR ground truth boxes due to occlusions, rolling shutter artifacts, and the limited range of the provided LiDAR scans (75m).
We apply an additional matching step by associating the 2D bounding boxes with our instance labels. First, we score matches between 2D boxes and instances by computing the IoU between each 2D box and the tightly-fitting bounding boxes around each instance mask, and then compute associations with bipartite matching. 

For boxes with existing 3D associations, we extend these tracks by propagating the existing ID to all other instances that match with the same 2D box. This resolves cases where only a object track misses 3D associations in a few frames. Then, we assign the remaining boxes without any matches to the ID of their corresponding 2D box, if any. 
Finally, to capture any additional cross-camera associations, we project all of the camera views onto the panorama, and associate instances which overlap in this joint representation.

In order to identify any instances that are still not associated with any ground truth boxes after these steps, we provide an additional mask for these instance pixels indicating that they are not tracked, similar to the \textit{crowd} mask used in single frame instance segmentation labels~\cite{Cordts16CVPR}.

\cutsectionup
\section{Benchmark and Evaluation Metrics} 
\cutsectiondown
In this section, we first describe the task of Panoramic Video Panoptic Segmentation (PVPS).
Then we review the evaluation metrics used in the literature, and propose a new metric designed for PVPS with an emphasis on consistent multi-object tracking and segmentation across multiple cameras.

\cutsubsectionup
\subsection{Problem Definition}
\cutsubsectiondown
We represent a multi-camera video sequence with $T$ frames and $M$ independent camera views as $\{\mathbf{I}^{1:T}_i\}_{i=1}^{M}$, where $\mathbf{I}_i^t$ is the $i$-th camera view captured at the $t$-th time step in the video sequence.
Along with the multi-view representation of the full scene, we define the panorama at $t$-th time step as $\mathbf{I}_\text{pano}^t$.
In the task of Panoramic Video Panoptic Segmentation (PVPS), we require a mapping $f$ of every pixel $(x, y, t, i)$ in the multi-camera video sequence to a semantic category $c \in \mathbf{C}$ and an instance ID $z$ consistent across camera views and temporal frames.
Here, $(x, y, t, i)$ indicates the spatial coordinate $(x, y)$ of the $i$-th camera view captured at the $t$-th time step, and $\mathbf{C}$ is the set of semantic categories.
Accordingly, we define the mappings $f_\text{id}$ and $f_\text{sem}$ for a particular instance ID $z$ and semantic category $c$ in~\equref{eqn:prob_inst_id} and~\equref{eqn:prob_sem_id}, respectively. The mapping functions are the building blocks of our proposed metric introduced in~\secref{sec:evaluation_metrics}.
\begin{align}
\label{eqn:prob_inst_id}
    f_\text{id}(z) &= \{(x,y,i,t) | f(x,y,i,t) = (c,z), c \in \mathbf{C}\}, \\
\label{eqn:prob_sem_id}    
    f_\text{sem}(c) &= \{(x,y,i,t) | f(x,y,i,t) = (c,*), c \in \mathbf{C}\}. 
\end{align}

Compared to the existing tasks including Video Panoptic Segmentaion (VPS) and Panoramic Semantic Segmentation, the proposed task is more challenging in the following aspects.
First, each individual camera has its own unique viewpoint and field-of-view such that the semantic class statistics are different across cameras (\eg, see~\figref{fig:class_dist_per_cam}).
This leads to a large domain gap between videos captured with different cameras.
Second, the instance ID prediction, with the long-term consistency across \textit{both} time and cameras, requires holistic scene understanding.

\cutsubsectionup
\subsection{Evaluation Metrics}
\label{sec:evaluation_metrics}
\cutsubsectiondown
In this subsection, we overview the existing Video Panoptic Segmentation (VPS) metric: Segmentation and Tracking Quality (STQ)~\cite{Weber2021NEURIPSDATA}, which we extend to evaluate the Panoramic Video Panoptic Segmentation (PVPS) task.

{\bf VPS Metric}\quad
We use $f$ and $g$ to indicate the prediction and ground-truth mapping, respectively.
We define the true positive associations (TPA)~\cite{Luiten20IJCV} of a specific instance as
$\text{TPA}(z_f,z_g) = |f_\text{id}(z_f) \cap g_\text{id}(z_g)|$, where $z_f$ is the predicted instance, $z_g \in \mathbf{G}$ is the ground-truth instance, and $\mathbf{G}$ is the set containing all unique ground-truth instances across cameras and temporal frames.
Similarly, false negative associations (FNA) and false positive associations (FPA) can be defined to compute the Intersection over Union ($\text{IoU}_\text{id}$) for evaluating tracking quality.
Formally, STQ is defined as follows.
\begin{align}\label{eqn:stq}
    STQ &= (AQ \times SQ)^{\frac{1}{2}},\\
    AQ &= \frac{1}{|\mathbf{G}|} \sum_{z_g \in \mathbf{G}} \frac{1}{|g_\text{id}(z_g)|}\sum_{z_f, |z_f \cap z_g| \neq \emptyset } \text{TPA}(z_f, z_g) \times \text{IoU}_\text{id}(z_f, z_g),\nonumber\\
    SQ &= \frac{1}{|\mathbf{C}|} \sum_{c \in \mathbf{C}} \frac{f_\text{sem}(c) \cap g_\text{sem}(c)}{f_\text{sem}(c) \cup g_\text{sem}(c)}.\nonumber
\end{align}

As defined in Eq~\eqref{eqn:stq}, STQ fairly balances segmentation and tracking performance, and is suitable for evaluating video sequences of arbitrary length. The Association Quality (AQ) measures the association quality for tracking classes, while the Segmentation Quality (SQ) measures the segmentation quality for semantic classes.
Specifically, AQ involves the $\text{IoU}_\text{id}$ computation for predicted instance IDs (and further weighted by true positive associations to encourage long-term tracking~\cite{Weber2021NEURIPSDATA}), while SQ is the typical semantic segmentation metric~\cite{Everingham10IJCV} (\ie, mean $\text{IoU}_\text{sem}$ for predicted semantic classes). 

\begin{figure}[!t]
    \centering
    \includegraphics[width=0.95\textwidth]{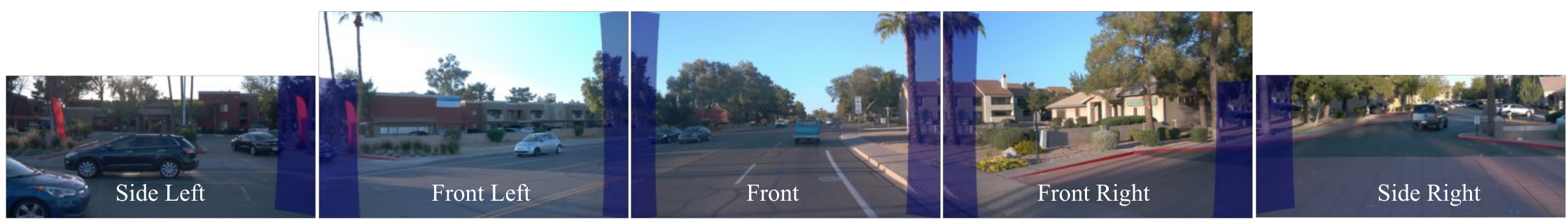}
    \cutcaptionup
    \caption{Visualization of the weights tensor for all cameras. Pixels in the blue region have weights $0.5$ during evaluation, as they are covered by two cameras.
    }
    \label{fig:weight_tensor}
    \cutcaptiondown
\end{figure}

{\bf PVPS Metric}\quad
We propose to extend the metric STQ~\cite{Weber2021NEURIPSDATA} for Panoramic Video Panoptic Segmentation (PVPS).
However, na\"ively adopting STQ for the multi-camera scenario results in a potential issue, where pixels in the overlapping regions covered by multiple cameras will be counted multiple times. 
Instead, we employ a simple and effective solution by exploiting the pixel-centric property of STQ. In particular, we weight each pixel prediction \wrt its coverage by the number of cameras, as determined by the mapping between the camera images and the panorama image. For example, if a pixel is covered by $N$ cameras (in our dataset, $N=2$), its prediction will contribute $1/N$ when computing AQ, and SQ.
We name the resulting metrics as \textit{weighted} STQ (wSTQ), since each pixel prediction takes a different weight depending on its coverage by the number of cameras. In \figref{fig:weight_tensor}, we visualize the weights for an example of five-camera images.

{\bf PS Metric}\quad
We also briefly review the metric PQ (panoptic quality)~\cite{Kirillov19CVPR} for evaluating image Panoptic Segmentation (PS), since we will build image-level baselines purely trained with image panoptic annotations.

For a particular semantic class $c$, the sets of true positives ($\text{TP}_\text{c}$), false positives ($\text{FP}_\text{c}$), and false negatives ($\text{FN}_\text{c}$) are formed by matching predictions $z_f$ to the ground-truth masks $z_g$ based on the IoU scores. A minimal threshold of greater than 0.5 IoU is chosen to guarantee unique matching. Formally,
\begin{align}
    \label{eq:pq}
    PQ_c &= \frac{\sum_{(z_f, z_g) \in \text{TP}_c } \text{IoU} (z_f, z_g)}{|\text{TP}_c| + \frac{1}{2} |\text{FP}_c| + \frac{1}{2} |\text{FN}_c|},
\end{align}
where the final PQ is then obtained by averaging $\text{PQ}_\text{c}$ over semantic classes.

\cutsectionup
\section{Experimental Results}
\cutsectiondown
In this section, we introduce our PVPS baselines, which exploit the property of multi-camera images by taking as input either individual camera views or panorama images (generated from all camera views). We then provide extensive experiments on the proposed dataset and metric.

\cutsubsectionup
\subsection{ViP-DeepLab Extensions as PVPS baselines}
\cutsubsectiondown
To tackle the new challenging PVPS task, we extend the state-of-art video panoptic segmentation method, ViP-Deeplab~\cite{Qiao2021CVPR}, to panoramic views.

{\bf Baseline Overview}\quad
For completeness, we first briefly review ViP-DeepLab~\cite{Qiao2021CVPR}. ViP-DeepLab extends the state-of-art image panoptic segmentation model, Panoptic-DeepLab~\cite{Cheng20CVPR}, to the video domain. Panoptic-DeepLab employs two separate prediction branches for semantic segmentation~\cite{Chen2018TPAMI} and instance segmentation~\cite{Kendall2018CVPR}, respectively. Both segmentation results are then merged~\cite{Yang2019Arxiv} to form the final panoptic segmentation result. To perform video panoptic segmentation, ViP-DeepLab adopts a two-frame image panoptic segmentation framework.
Specifically, during training, ViP-DeepLab takes a pair of image frames as input and their panoptic segmentation ground-truths as training target. During inference, ViP-DeepLab performs two-frame image panoptic predictions at each time step, and continues the inference process for every two consecutive frames (\ie, with one overlapping frame at the next time step) in a video sequence.
The predictions in the overlapping frames are ``stitched" together by propagating instance IDs based on mask IoU between region pairs (\ie, if two masks have high IoU overlap, they will be re-assigned with the same instance ID), and thus temporally consistent IDs are obtained (see Fig. 4 of Qiao \etal~\cite{Qiao2021CVPR} for an illustration). We refer this post-processing as ``panoptic stitching over time".

\begin{figure}[!t]
     \centering
    \includegraphics[width=0.95\textwidth]{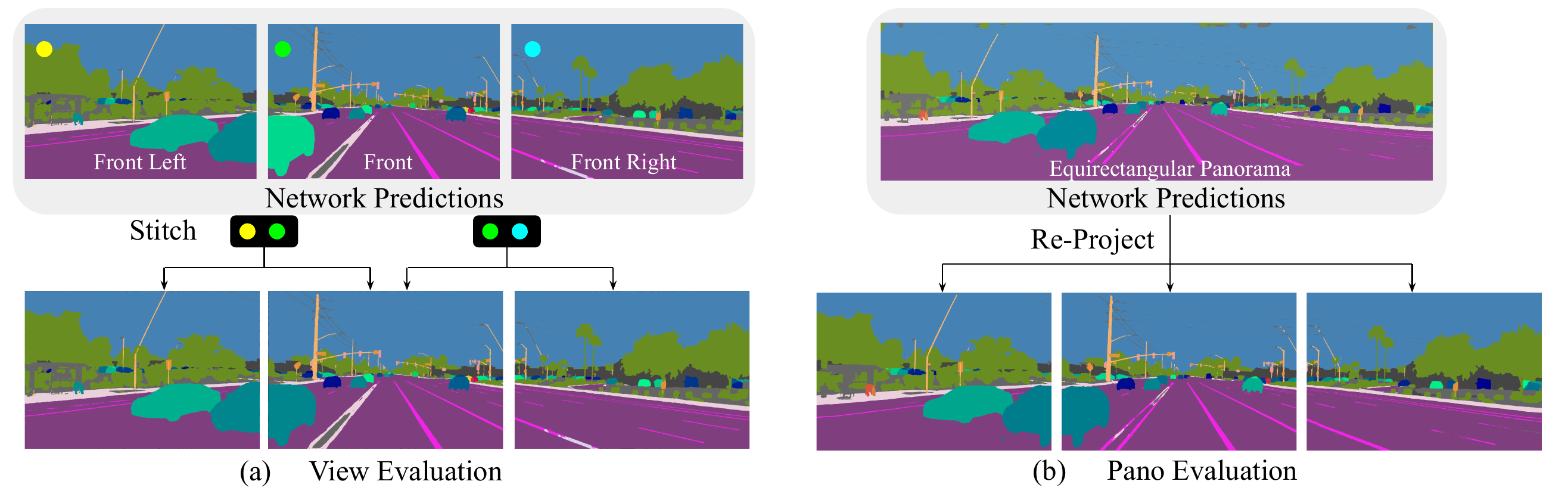} 
    \cutcaptionup
    \caption{We experiment with two evaluation schemes: (a) View and (b) Pano. The View evaluation scheme takes individual camera views as input and generates their panoptic predictions, which are then ``stitched over cameras" to obtain consistent instance IDs between cameras. The Pano evaluation scheme takes as panorama images as input and generates panoramic panoptic predictions, which are then reprojected back to each camera for evaluation.}
    \label{fig:evaluation_scheme}
    \cutcaptiondown
\end{figure}

{\bf Baseline Extension for PVPS}\quad
We explore several ViP-DeepLab extensions for PVPS, which  takes as input individual camera views or panorama images (generated from all camera views).
The input types could be different during training and evaluation.
Specifically, we define three training schemes: \textit{View}, \textit{Pano}, and \textit{Ensemble-View}.
The View scheme refers to the case where ViP-DeepLab is trained with images from all camera views, while Pano means the model is trained with full panorama images.
The Ensemble-View scheme refers to the case where we have five camera-specific ViP-DeepLab models, each of which is trained and evaluated on their own camera images. We also have two evaluation schemes: \textit{View} and \textit{Pano}. The View scheme refers to the case where the trained model is fed with images from individual camera views and generates the corresponding panoptic predictions for each view.
However, the predicted instance IDs are not consistent between cameras, since the predictions are made independently for each view.
To generate consistent instance IDs between cameras, we propose a similar method to ``panoptic stitching over time": if two masks have high IoU overlap in the overlapping regions between two cameras' field-of-view, we re-assign the same instance ID for them, resulting in the ``panoptic stitching over cameras" post-processing method.
For the Pano evaluation scheme, the model is fed with panorama images and generates panoramic panoptic predictions. We then re-project panoramic panoptic predictions onto each camera for evaluation. Note that, for the Pano evaluation scheme, the instance IDs are consistent between cameras by nature. We visualize the evaluation schemes in~\figref{fig:evaluation_scheme}.

{\bf Implementation Details}\quad
We build our image-based and video-based baselines on top of Panoptic-DeepLab~\cite{Cheng20CVPR} and ViP-DeepLab~\cite{Qiao2021CVPR}, respectively, using the official code-base~\cite{deeplab2_2021}.
The training strategy follows Panoptic-DeepLab and ViP-DeepLab. Specifically, the models are trained with 32 TPU cores for 60k steps, batch size 32, Adam~\cite{kingma2014adam} optimizer and a poly schedule learning rate of $2.5 \times 10^{-4}$. We use an ImageNet-1K-pretrained~\cite{Russakovsky15IJCV} ResNet-50~\cite{He2016CVPR} with stride 16 as the backbone (using atrous convolution~\cite{Chen2015ICLR}).
For image-based methods, we use the crop size $1281 \times 1921$ during training, while, during inference, we use the whole image (or panorama).
We use a similar strategy for the video-based methods, but we use a ResNet-50 backbone with stride 32 and crop size $641\times 961$ due to memory constraints.

\cutsubsectionup
\subsection{Qualitative Evaluation}
\begin{figure}[!t]
    \centering
    \includegraphics[width=0.9\textwidth]{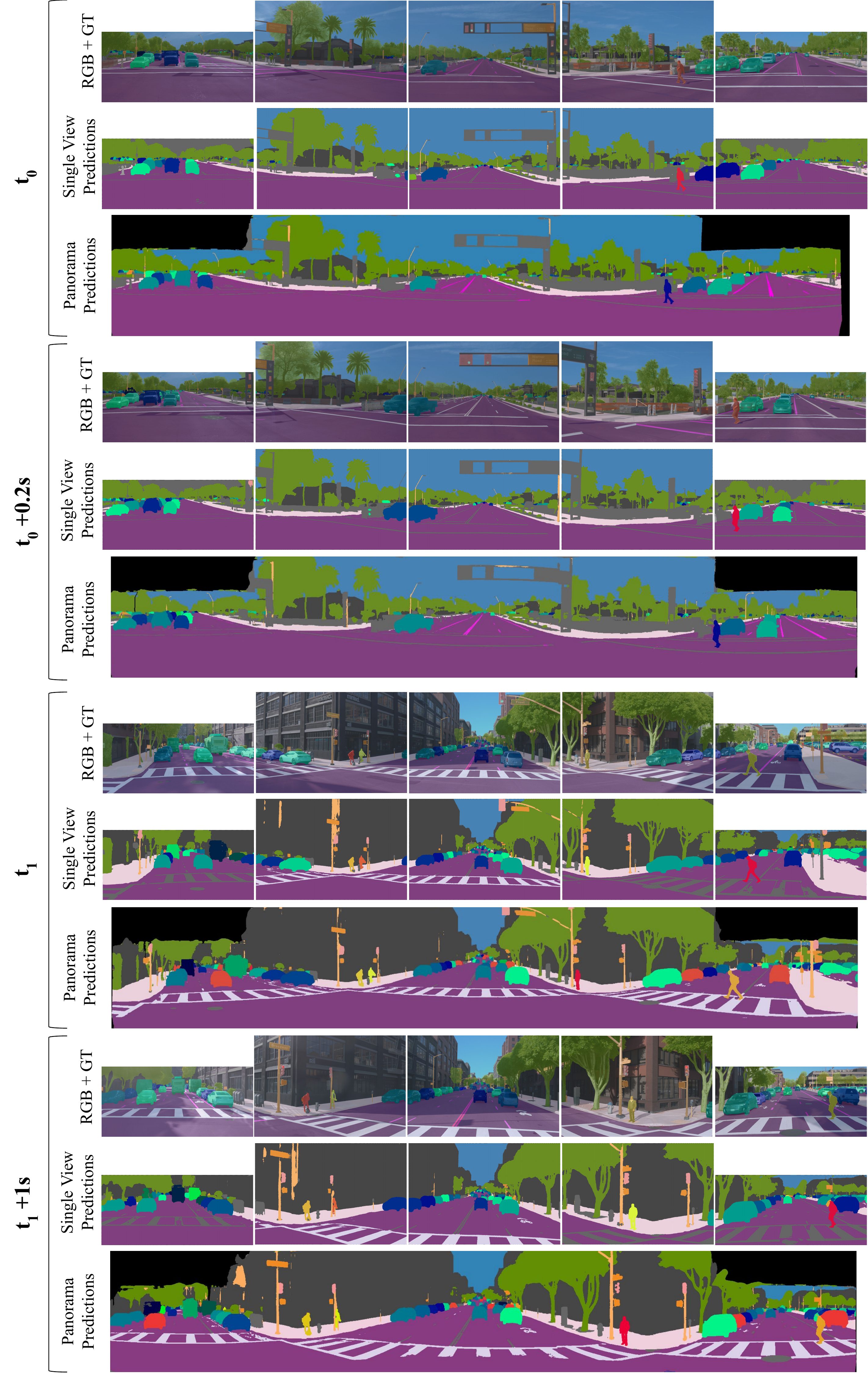}
    \caption{Comparison of qualitative results from our baseline ViP-DeepLab~\cite{Qiao2021CVPR} models over different time intervals. Results show models trained on single images with panoptic stitching over cameras, and trained directly on panorama images. Our baseline models show strong performance for the majority of the scene, although tracking small/distant objects and crowded scenes remains challenging.
    }
    \label{fig:example_prediction}
    \cutcaptiondown
\end{figure}

\cutsubsectiondown
In ~\figref{fig:example_prediction}, we provide qualitative results from our two ViP-DeepLab baselines, both trained and evaluated on single images and on panorama images (\ie, one model uses View schemes for both training and evaluation, and the other uses Pano schemes for both training and evaluation), over two (non-adjacent) temporal frames. From these results, we can see that the baseline models are able to accurately track objects in very dense scenes. In addition, we note that there are some qualitative benefits provided by the panorama model in these examples. In particular, the single view model has an inconsistent prediction on the crosswalk in the left and right images for the single view model, but the panorama model is able to attain the full context of the scene and avoids this mistake. In addition, the single view models fail to track the car crossing the front right and side right cameras at $t_0$, but the panorama is again able to track this object correctly.

\cutsubsectionup
\subsection{Baseline Comparisons}
\cutsubsectiondown

{\bf Video-based Baselines}\quad
In~\tabref{tab:image_and_video_baseline}(a), we provide video-based baseline comparisons using ViP-DeepLab~\cite{Qiao2021CVPR}, evaluated by the proposed weighted STQ (wSTQ). We compare different training and evaluation schemes. As shown in the table, when both evaluated with View scheme, training with View scheme performs better than training with Ensemble-View by 0.86\% wSTQ. That is, training a single model with all the camera views performs better than training five camera-specific models with its own camera views. Also, when training with View scheme, using the Pano evaluation scheme degrades the performance by 2.91\% wSTQ. When training with Pano scheme, using View scheme is better than Pano scheme for evaluation. We think it is caused by the asymmetry between training and evaluation settings. We could not use whole panorama images with resolution $1000 \times 5875$  as input during training (due to memory limit), and thus we only use a smaller crop size $641 \times 961$. Ideally, the model should be evaluated with the same setting as its training. The current best setting is trained and evaluated with the View scheme, reaching 17.78\% wSTQ. We observe that our dataset is very challenging in terms of both tracking and segmentation, since our best wAQ is only 8.21\% and best wSQ is 39.78\%.

{\bf Image-based Baselines}\quad
In~\tabref{tab:image_and_video_baseline}(b), we provide image-based baseline comparisons using Panoptic-DeepLab~\cite{Cheng20CVPR}, evaluated by image panoptic segmentation metric PQ~\cite{Kirillov19CVPR} and semantic segmentation metric mIoU~\cite{Everingham10IJCV}. Basically, we observe the same trend of image-based baselines and video-based baselines.

\begin{table}[t]
    \cutcaptionup
    \caption{Quantitative evaluation: during training and evaluation, the baselines can take different types of inputs: 
    View: individual camera views; Pano: panoramas; Ensemble-View: camera-specific views. Results include (a) video-baseline comparison using ViP-DeepLab, measured by weighted Segmentation and Tracking Quality (wSTQ); and (b) image-baseline comparison using Panoptic-DeepLab, measured by Panoptic Quality (PQ) and mean Intersection-over-Union (mIoU). }
    \label{tab:image_and_video_baseline}
    \begin{subtable}[h]{0.515\textwidth}
    \cuthalftablecaptionup
    \caption{Video-baseline Comparison}
    \cuthalftablecaptiondown
    \centering
    \resizebox{\textwidth}{!}{
    \begin{tabular}{c|c|c|c|c}
    \toprule
    Training Scheme & Eval Scheme & wSTQ & wAQ & wSQ \\
    \midrule
    Ensemble-View & View & 16.92 & 7.61 & 37.33 \\
    View & View & \textbf{17.78} & \textbf{8.21} & 38.46 \\
    View & Pano & 14.87 & 6.13 & 36.04 \\
    Pano & View & 17.56 & 8.11 & 38.04 \\
    Pano & Pano & 15.72 & 6.22 & \textbf{39.78} \\
    \bottomrule
    \end{tabular}
    } 
    \label{tab:video_baselines}
    \end{subtable}
    \hfill
    \begin{subtable}[h]{0.45\textwidth}
    \cuthalftablecaptionup
    \caption{Image-baseline Comparison}
    \cuthalftablecaptiondown
    \centering
    \resizebox{\textwidth}{!}{
    \begin{tabular}{c|c|c|c}
    \toprule
    Training Scheme & Eval Scheme & PQ & mIoU \\
    \midrule
    Ensemble-View & View & 35.70 & 48.15 \\
    View & View & \textbf{40.00} & \textbf{53.64} \\
    View & Pano & 33.65 & 50.61 \\
    Pano & View  & 38.93 & 51.65 \\
    Pano & Pano  & 36.32 & 52.19 \\
    \bottomrule
    \end{tabular}
    }
    \label{tab:panoptic_deeplab}
    \end{subtable}
\end{table}

\cutsubsectionup
\subsection{Ablation Studies}
\cutsubsectiondown

{\bf Transferability of Models between Viewpoints}\quad
In this ablation, we measure the ability to transfer models trained on one viewpoint to a different viewpoint. As shown in~\tabref{tab:video_based_transferability}, we have the following observations: First, all models, even trained on left side views, perform better on right side views. This phenomenon is due to the ego-vehicle driving on the right side of the road, and providing wider scope, more instances, and smaller objects on the left side (\ie, the left side views are more challenging). Second, the front camera performance is inferior compared to the other cameras.
We hypothesize that the front camera captures more diverse and challenging views, \eg, vehicles driving in multiple directions, more dynamic and smaller objects, making tracking more challenging.

\begin{table}[!t]
\cutcaptionup
\caption{View transferability on our video-based baselines, measured by wSTQ. We evaluate models (1st column) trained on a specific view \wrt other camera views. The last row, MultiCamera, refers to the model trained with all camera views (\ie, training scheme View), and the last column, All, denotes the evaluation set using all camera views (\ie, evaluation scheme View).
}

\centering
\scriptsize
\begin{tabular}{l|c|c|c|c|c|c}
\toprule
Model $\setminus$ Eval  & Side Left & Front Left & Front & Front Right & Side Right & All \\
\midrule
Side Left & \textbf{18.79} & 17.41 & 14.56 & 19.06 & 19.40 & \textbf{16.31} \\
Front Left & 16.88 & \textbf{18.39} & 12.84 & 19.22 & 18.49 & 15.36 \\
Front & 16.58 & 18.02 & 14.54 & 18.55 & 18.96 & 15.98 \\
Front Right & 16.56 & 17.36 & \textbf{14.99} & \textbf{19.40} & 19.16 & 16.18 \\
Side Right & 17.91 & 16.50 & 13.16 & 18.23 & \textbf{20.47} & 15.65 \\
\midrule
MultiCamera & 20.11 & 19.54 & 15.63 & 20.67 & 21.53 & 17.78 \\
\bottomrule
\end{tabular}
\label{tab:video_based_transferability}
\cutcaptiondown
\end{table}

\cutsectionup
\section{Conclusion}
\cutsectiondown
In this work, we presented a new benchmark, the Waymo Open Dataset: Panoramic Video Panoptic Segmentation (\ourdataset) dataset. Our benchmark extends video panoptic segmentation to a more challenging multi-camera setting that requires consistent instance IDs both across cameras and over time. Our dataset is an order of magnitude larger than all the existing video panoptic segmentation datasets. We establish several strong baselines evaluated with a new metric, wSTQ, that takes multi-camera, multi-object tracking and segmentation into consideration.
We will make our benchmark publicly available, and we hope that it will facilitate future research on panoramic video panoptic segmentation.

\clearpage

\bibliographystyle{splncs04}
\bibliography{egbib}
\clearpage

\end{document}